\documentclass{article}
\usepackage{iclr2027_conference,times}

\usepackage{amsmath,amsfonts,bm}

\def\eqref#1{equation~\ref{#1}}

\def\1{\bm{1}}

\DeclareMathAlphabet{\mathsfit}{\encodingdefault}{\sfdefault}{m}{sl}
\SetMathAlphabet{\mathsfit}{bold}{\encodingdefault}{\sfdefault}{bx}{n}

\usepackage{caption}
\usepackage{hyperref}
\usepackage{url}
\usepackage{graphicx}
\usepackage{booktabs}
\usepackage{multirow}
\usepackage{amsmath}
\usepackage{amssymb}
\usepackage{wrapfig}
\usepackage[table]{xcolor}
\usepackage{algorithm}
\usepackage{algorithmic}

\title{TERRA: Learning Transportable Latent Actions through Temporal Effect Representation and Relational Alignment}

\author{
Tianxingjian Ding$^{1}$  \quad Mubarak Shah$^{1,\dagger}$ \quad Yu Tian$^{1,\dagger}$\\
{\normalfont $^{1}$Institute of Artificial Intelligence, University of Central Florida} \\
}

\iclrfinalcopy

\begin{document}
\maketitle
\lhead{Preprint}
{\renewcommand{\thefootnote}{\fnsymbol{footnote}}\footnotetext[2]{Corresponding authors.}}

\begin{abstract}
Latent actions supervise robot policies with action-like codes inferred
from visual transitions, and their usefulness hinges on two questions:
what a code keeps from a transition, and whether it still means the same
thing when reused in a different initial state. The first is a tension
in time: an endpoint difference discards how motion unfolds, while the
full sequence admits nuisance variation. The second is left open by
reconstruction, which only ever observes a latent together with the
state it came from. We argue that both questions can be answered in the
same place. \textbf{TERRA} (\emph{Temporal Effect Representation and
Relational Alignment}) describes a transition by a compact temporal effect, its net
feature change together with a low-order within-window dynamics
component, and learns a continuous latent from this effect. The same
effect space then serves as the reference for reuse: \textbf{Effect-Anchored
Transport (EAT)} decodes a latent in other initial states and anchors the
resulting effect to the one observed at its source, so that the latent
is shaped by what it does across contexts rather than only by the
transition it came from. With frozen linear readers, TERRA predicts
actions more accurately than UniVLA and a LAPA-style baseline, degrades
more slowly under visual distractors, and keeps transported transitions
faithful to the donor action as the recipient context moves farther
away; a same-budget control shows that these gains come largely from EAT. At matched pretraining scale, the
complete system reaches 93.4\% average success on LIBERO, compared with
91.8\% for UniVLA.
\end{abstract}
\section{Introduction}
\label{sec:introduction}
Latent actions provide an intermediate supervision signal between visual
transitions and robot control. A latent-action model assigns an action-like
representation to an observed transition, which can then supervise a policy
conditioned on the current observation and instruction. Recent work has
explored discrete task-centric latents~\citep{ye2025lapa,bu2025univla},
continuous latent motion from frame-pair feature
differences~\citep{yang2026como}, and structured continuous action
spaces~\citep{li2026rotvla}. Despite this progress, a basic question remains
unresolved: \emph{what should a latent action preserve from a visual
transition, and how should it behave when reused?}
\vspace{-1mm}
\begin{figure}[t]
    \centering
    \includegraphics[width=\textwidth]{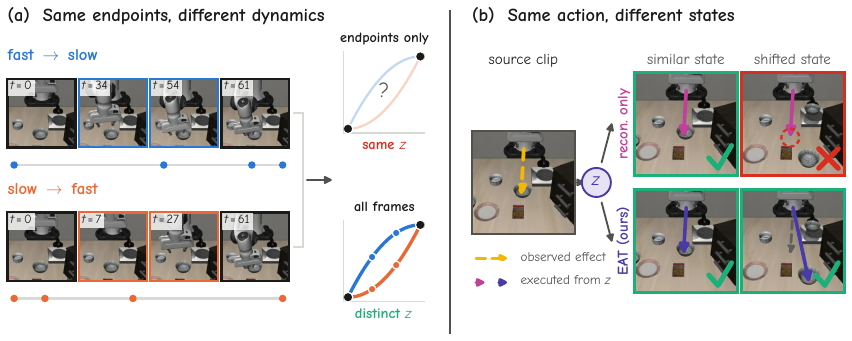}
    \caption{\textbf{Two questions for a latent action}.
(a) \emph{What to keep.} Two executions share their first and last frames
($t{=}0$, $t{=}61$) but move at different rhythms (fast-then-slow vs.\
slow-then-fast, dots mark frame times). A latent built from the endpoint
difference alone assigns them the same code; adding a within-window dynamics
component keeps them distinct.
(b) \emph{How to behave.} The latent $z$ of a source clip (yellow: its
observed effect, the gripper lowering onto the bowl) is decoded from two
other initial states. Arrows show the effect each model decodes. Trained by
reconstruction only, the effect follows the source in a similar state but
drifts once the bowl moves; with EAT it stays consistent with the source
effect in both, so the same latent transfers across states.}
\vspace{-1mm}
    \label{fig:teaser}
\end{figure}

A first challenge is \textbf{selective temporal information preservation}.
Strong compression can remove distinctions that matter for control, whereas
increasing latent capacity can also admit action-irrelevant appearance and
exogenous variation~\citep{yang2026como,lee2026why}. This tension is
particularly important in time. Continuous latent-motion models operate on a frame pair, so their input
reduces to the endpoint difference: it records how much the scene changed,
not how the change unfolded within the window. Two executions that reach the
same state at different speeds therefore receive the same latent
(Fig.~\ref{fig:teaser}a), although they correspond to different actions.
Conversely, retaining the entire temporal sequence exposes the representation
to substantially more visual variation.

A second challenge is \textbf{cross-context effect consistency}. Standard
reconstruction objectives observe only matched state--transition pairs and
can therefore encode context-specific variation into the
latent~\citep{lee2026why}. Policy supervision is most useful when similar effects receive compatible
latents across episodes, so it matters how a latent behaves when applied in a
different initial state; we refer to this reuse as \emph{transport}. A latent
that fits its source transition can still decode to a different effect once
the scene changes (Fig.~\ref{fig:teaser}b).
 
\begin{figure*}[t]
    \centering
    \includegraphics[width=\textwidth]{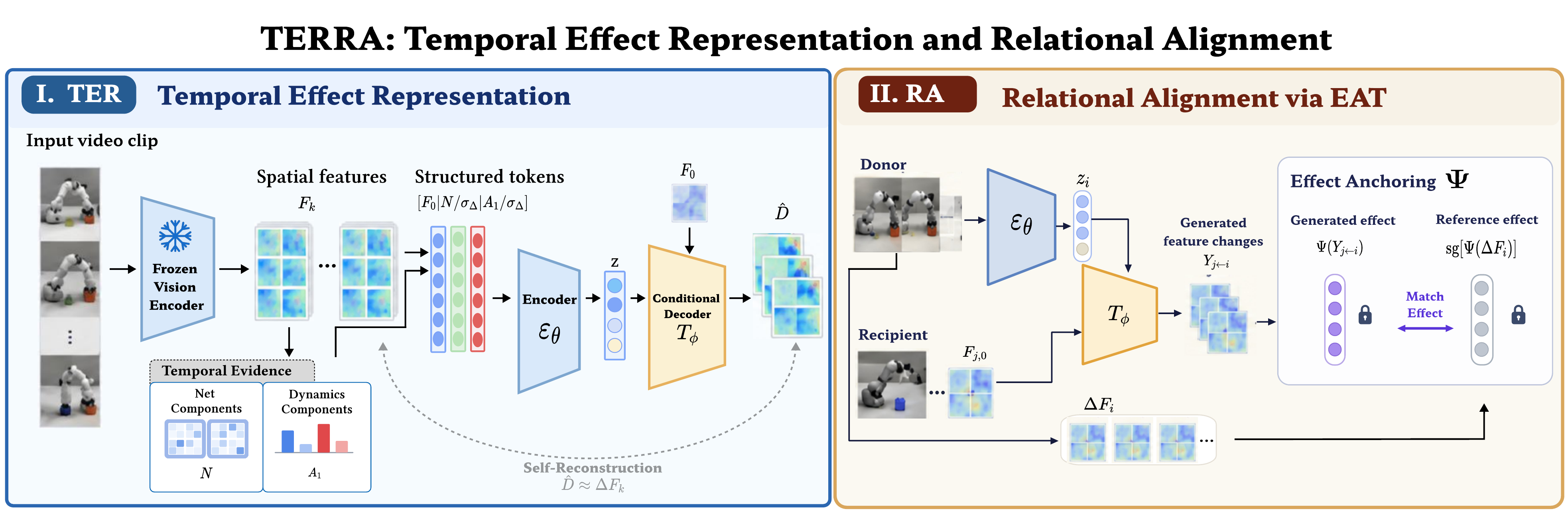}
    \caption{
    \textbf{Overview of TERRA.}
    Temporal Effect Representation compresses the initial-state feature $F_0$
    and the temporal evidence (net component $N$ and dynamics component $A_1$)
    into a continuous latent $z$. Effect-Anchored Transport decodes a donor
    latent in recipient states and anchors the resulting effect to the donor's
    observed effect. See Sec.~\ref{sec:method} for details.
    }
    \label{fig:terra}
\end{figure*}

We address both challenges with one principle: \emph{the compact effect space
that decides what a latent should keep also provides the reference for how it
should behave in a new context}. We instantiate this principle with
\textbf{TERRA} (\emph{Temporal Effect Representation and Relational
Alignment}), which first summarizes a short transition by its \emph{temporal
evidence}, a net component that captures accumulated feature change and a
dynamics component that captures its low-order within-window trend, and
compresses this evidence into a continuous latent. It then applies
\textbf{Effect-Anchored Transport (EAT)}: the latent is decoded in recipient
states from the same data source, and the direction of the resulting effect
is anchored to the one observed at the donor. Relative-effect consistency,
transition-distribution matching, and nuisance invariance regularize this
refinement. Figure~\ref{fig:terra} summarizes the framework.
 
Adding the dynamics component to the endpoint difference lowers action NMSE
and makes the within-window action trend linearly recoverable. EAT lowers
action NMSE further and keeps transported transitions faithful to the donor
action as recipient distance grows, which the same amount of additional
reconstruction does not achieve. The complete system reaches 93.4\% average
LIBERO success against 91.8\% for UniVLA at matched pretraining scale.

Our contributions are:
\begin{itemize}
    \item A \textbf{temporal-effect representation} that extends frame-pair
    latent motion with a low-order within-window dynamics component,
    retaining how change unfolds without encoding the full sequence.
 
    \item \textbf{Effect-Anchored Transport}, which uses the same effect space
    to supervise how a latent behaves when decoded in other initial states,
    constraining what the latent does rather than what it is.
 
    \item Controlled evaluations showing that each component contributes to
    action alignment, nuisance robustness, and transport stability, and a
    matched-scale system comparison showing higher LIBERO success than
    UniVLA.
\end{itemize}

\section{Related Work}

\label{sec:related_work}
\paragraph{Latent action representations.}
Latent action models infer action-like representations from visual
transitions and use them as intermediate supervision for robot learning.
Early work learned inverse-dynamics labels from unlabeled video~\citep{baker2022vpt}
and discrete latent actions for controllable generation and
policies~\citep{bruce2024genie,schmidt2024lapo}; LAPA and UniVLA scale this
recipe to vision-language-action pretraining with quantized
codes~\citep{ye2025lapa,bu2025univla}, and Moto, IGOR, AdaWorld, and villa-X
refine what such codes capture~\citep{chen2025moto,chen2024igor,gao2025adaworld,chen2025villax}.
More recent work explores continuous representations to preserve richer motion
structure: CoMo learns continuous latent motion from frame-pair feature
differences~\citep{yang2026como}, and RotVLA introduces structured continuous
latents with temporal composition for VLA learning~\citep{li2026rotvla}.
Zhang et al.~\citep{zhang2025lamlearn} analyze what these models actually
encode. These works establish both discrete and continuous latent actions as
viable interfaces between visual transitions and downstream control.
\paragraph{Cross-context grounding and latent consistency.}
A separate line of work studies how latent actions remain meaningful beyond
the transition from which they are inferred.
Olaf-World orients latent actions using temporal feature differences from a
frozen video encoder for controllable video world models
\citep{jiang2026olaf}, while cycle-based approaches regularize latent identity
through generated transitions \citep{chen2025vivla}.
LAOM shows that action-correlated distractors degrade latent actions learned
by reconstruction and measures latent quality by linear
probing~\citep{nikulin2025laom}; recent analyses further examine nuisance
sensitivity, action recoverability, and the relationship between latent-quality
proxies and downstream control~\citep{lee2026why,bu2026whatmatters}.
\paragraph{Continuous VLA prediction.}
Continuous action prediction is increasingly common in vision-language-action
models.
OpenVLA-OFT replaces autoregressive action-token decoding with continuous
action heads \citep{kim2025oft}, while $\pi_0$ uses a flow-matching action
expert \citep{black2025pi0}.
RotVLA jointly models continuous latent and physical actions
\citep{li2026rotvla}, and LAFP studies latent-policy learning with flow matching
\citep{lyu2026lafp}.

These prior works have studied temporal latent construction, cross-context consistency, and continuous policy prediction largely as separate problems. TERRA uses one effect space for both: temporal evidence defines what the latent keeps, and Effect-Anchored Transport uses the same evidence to supervise how it behaves in other states.

\section{TERRA: Temporal Effect Representation and Relational Alignment}
\label{sec:method}

TERRA learns transportable continuous latent actions through two
complementary components: \emph{Temporal Effect Representation} (TER)
and \emph{Relational Alignment} (RA).
TER constructs a compact latent from the temporal evidence of an
observed transition, while RA refines how this latent behaves when
applied in a different initial state, through Effect-Anchored
Transport (EAT).

We call a latent \emph{transportable} within a data source if, when
applied in a different initial state, the resulting transition preserves
the direction of its source temporal effects.
We refer to this cross-context reuse as \emph{transport}.
Both components use frozen visual features without robot action labels.
After refinement, the latent tokenizer is frozen and used to provide
latent-action targets for downstream policy training.

\subsection{Temporal Effect Representation (TER)}
\label{sec:temporal_effect_representation}

\paragraph{Temporal evidence.}
For a transition
$x_i=(o_{i,0},\ldots,o_{i,K})$,
we extract frozen DINOv2 features~\citep{oquab2023dinov2} and spatially
pool them into
$F_{i,k}\in\mathbb{R}^{S\times C}$,
where $S$ denotes the number of spatial cells and $C$ the feature
dimension.
Let
$\Delta F_{i,k}=F_{i,k+1}-F_{i,k}$.
We omit the transition index $i$ when unambiguous.

Rather than retaining either only the endpoint difference or the full
increment sequence, TERRA summarizes a transition by two low-order
temporal components, which we call its \emph{temporal evidence}.
For any increment sequence
$Y=(Y_0,\ldots,Y_{K-1})$,
we define the \emph{net component} $N$ and the \emph{dynamics component}
$A_1$ as
\begin{equation}
N(Y)
=
\sum_{k=0}^{K-1}Y_k,
\qquad
A_1(Y)
=
\sum_{k=0}^{K-1} b_1(k)\,Y_k,
\qquad
\langle b_1,\mathbf{1}\rangle=0 .
\label{eq:temporal_effects}
\end{equation}
The net component corresponds to the constant temporal mode and captures
accumulated change; for observed transitions,
$N(\Delta F)=F_K-F_0$ by telescoping.
The dynamics component uses the lowest-frequency nonconstant DCT-II mode
$b_1$ and captures coarse within-window variation.
Together they retain the two lowest-order temporal components of the
increment sequence.
The temporal evidence is the measured form of the temporal effect that
TERRA represents; Section~\ref{sec:eat} reuses the same two components
to supervise transport.

We collect the temporal evidence into the \emph{effect vector}
\begin{equation}
\xi(Y)
=
\operatorname{vec}
\left[
N(Y),A_1(Y)
\right].
\label{eq:effect_vector}
\end{equation}

\paragraph{Structured tokens and encoding.}
Feature increments are normalized by a single global scale
$\sigma_\Delta$, defined as the root-mean-square magnitude of raw
DINOv2 feature increments estimated once before training and then kept
fixed.

We concatenate the initial-state feature and the normalized temporal
evidence along the token dimension into the \emph{structured tokens}
\[
\left[
F_0
\mid
N(\Delta F)/\sigma_\Delta
\mid
A_1(\Delta F)/\sigma_\Delta
\right].
\]
A transformer with spatial and evidence-type embeddings aggregates the
three token groups through four learned queries.
Their outputs form
\[
z
=
E_\theta(x)
\in
\mathbb{R}^{4\times64},
\]
with no quantization or predefined spatial or temporal roles for the
latent slots.
We refer to $E_\theta$ as the \emph{latent tokenizer}.

\paragraph{Self-reconstruction.}
The conditional decoder $T_\phi$ maps the initial-state feature and the
latent to a normalized prediction of the increment sequence,
\begin{equation}
\hat{D}
=
T_\phi(F_0,z)
\in
\mathbb{R}^{K\times S\times C}.
\label{eq:decoder}
\end{equation}
The tokenizer and decoder are jointly trained so that
$\hat{D}\approx\Delta F/\sigma_\Delta$:
\begin{equation}
\mathcal{L}_{\mathrm{self}}
=
\mathbb{E}_{x}
\left[
\frac{1}{KSC}
\left\|
\hat{D}
-
\frac{\Delta F}{\sigma_\Delta}
\right\|_F^2
\right],
\label{eq:self_reconstruction}
\end{equation}
where $\Delta F$ stacks all $K$ increments.
Predicting the increment sequence requires the decoder to model how
change is distributed within the window rather than only its endpoint.
The tokenizer itself, however, receives only the temporal evidence
$N$ and $A_1$; higher-order temporal detail is not available to it as
input.

\subsection{Relational Alignment (RA) via Effect-Anchored Transport}
\label{sec:eat}

The representation above is learned only from matched
state--transition pairs.
Such reconstruction does not determine how the same latent should behave
when applied in a different initial state.
By the definition above, the desired behavior is that the transported
transition preserve the donor's temporal-effect direction;
EAT enforces this behavior directly.

We sample donor transitions and recipient initial states from the same
data source, limiting large embodiment and observation-domain shifts.
EAT combines four complementary objectives:
\emph{effect anchoring},
\emph{relative-effect consistency},
\emph{transition-distribution matching}, and
\emph{nuisance invariance}.

\paragraph{Transport.}
Let $x_i$ be a donor transition and $F_{j,0}$ a recipient initial state
from another episode of the same source.
We encode the donor and decode its latent in the recipient state,
\begin{equation}
z_i
=
E_\theta(x_i),
\qquad
\hat{D}_{j\leftarrow i}
=
T_\phi(F_{j,0},z_i),
\qquad
Y_{j\leftarrow i}
=
\sigma_\Delta\,
\hat{D}_{j\leftarrow i}.
\label{eq:transport}
\end{equation}
Here $Y_{j\leftarrow i}$ is the generated feature-increment sequence
induced by donor latent $z_i$ in recipient state $F_{j,0}$, expressed in
feature-increment units so that the temporal evidence of
Eq.~\ref{eq:temporal_effects} can be evaluated on it.
The tokenizer sees only the donor transition, and the decoder receives
no recipient future frames.
EAT introduces no additional trainable transport network:
$T_\phi$ is the same conditional decoder used for self-reconstruction.

\paragraph{Effect anchoring.}
We compare effects through a fixed \emph{effect descriptor}
\begin{equation}
\Psi(Y)=\operatorname{std}\bigl(R^{\top}\xi(Y)\bigr),
\label{eq:effect_descriptor}
\end{equation}
where $R$ is a fixed Gaussian random projection and $\operatorname{std}$ standardizes each coordinate
with mean and variance estimated once from real transitions
(Appendix~\ref{app:eat_details}).
$\Psi$ has no learnable parameters, so the temporal evidence used to
construct the representation is the same quantity used to evaluate its
behavior after transport.
The observed donor transition provides the reference:
\begin{equation}
\mathcal{L}_{\mathrm{effect}}
=
\mathbb{E}_{i,j}
\left[
1-
\cos
\left(
\Psi(Y_{j\leftarrow i}),
\operatorname{sg}
[
\Psi(\Delta F_i)
]
\right)
\right],
\label{eq:effect_anchor}
\end{equation}
where $\operatorname{sg}$ denotes stop-gradient.
This objective aligns the direction of the transported effect with that
of the observed donor transition while leaving its magnitude free to
depend on context.
The generated branch remains differentiable, so
$\mathcal{L}_{\mathrm{effect}}$ updates both $T_\phi$ and the online
tokenizer $E_\theta$.

\paragraph{Relative-effect consistency.}
Effect anchoring constrains individual transported effects.
We additionally encourage relationships between different latent effects
to remain stable across recipient contexts. For this objective, we maintain an exponential-moving-average target
tokenizer $E_{\bar{\theta}}$ that tracks $E_\theta$ but receives no
gradients.
Detached target latents from recent same-source samples provide the donor
latents for relative-effect consistency.
Let $\bar z_i$ and $\bar z_{i'}$ denote two such donor latents.
For recipient states $F_{j,0}$ and $F_{j',0}$, define
\begin{equation}
e_s^q
=
\Psi
\left(
\sigma_\Delta
T_\phi(F_{s,0},\bar z_q)
\right),
\qquad
q\in\{i,i'\},
\quad
s\in\{j,j'\},
\label{eq:transported_effect}
\end{equation}
and
\[
r_s^{i,i'}
=
e_s^i-e_s^{i'}.
\]
We optimize
\begin{equation}
\mathcal{L}_{\mathrm{rel}}
=
\mathbb{E}_{i,i',j,j'}
\left[
1-
\cos
\left(
r_j^{i,i'},
r_{j'}^{i,i'}
\right)
\right].
\label{eq:relative_effect}
\end{equation}

Effect anchoring constrains the direction of each transported effect,
whereas relative-effect consistency stabilizes how different latent
effects compare across recipient contexts.
Together, they reduce context dependence in the effect space measured by
$\Psi$ without requiring the full generated transition to be
state-independent.

Because the donor latents used by
$\mathcal{L}_{\mathrm{rel}}$ are detached EMA-target representations,
this term directly regularizes the shared decoder rather than
backpropagating through the donor tokenizer.
The online tokenizer is refined directly through
$\mathcal{L}_{\mathrm{self}}$,
$\mathcal{L}_{\mathrm{effect}}$, and
$\mathcal{L}_{\mathrm{inv}}$.

\paragraph{Transition-distribution matching.}
The temporal evidence does not constrain every component of the
generated transition.
We therefore use maximum mean discrepancy (MMD)~\citep{gretton2012kernel}
to match the empirical distributions of projected real and generated
feature increments,
denoted $\mathcal{L}_{\mathrm{dist}}$.
This term operates in transition space rather than latent space and
does not assign a ground-truth counterfactual transition to an individual
recipient.
Its exact form is given in Appendix~\ref{app:eat_details}.

\paragraph{Nuisance invariance.}
To suppress visual variation that should not determine the latent action, we
use a BYOL-style objective~\citep{grill2020byol}: the online tokenizer
receives a temporally consistent nuisance-perturbed view of a transition,
while the EMA target tokenizer receives the corresponding clean view.
The perturbation family consists of mild spatial cropping together with
brightness and contrast variation, shared across all frames in the clip.
We minimize the cosine discrepancy between the online and EMA-target
latent representations, denoted $\mathcal{L}_{\mathrm{inv}}$.

Importantly, these augmentations are used only for
$\mathcal{L}_{\mathrm{inv}}$.
The effect descriptor $\Psi$ and both effect-based objectives are
computed from unaugmented transitions.
Thus, the spatial perturbation used for latent invariance does not alter
the image-grid coordinates on which effect anchoring is defined.
Exact augmentation and target-network details are provided in
Appendix~\ref{app:eat_details}.

\paragraph{Joint optimization.}
We retain self-reconstruction on matched transitions throughout EAT
refinement and optimize
\begin{equation}
\mathcal{L}_{\mathrm{EAT}}
=
\mathcal{L}_{\mathrm{self}}
+
\lambda_e\mathcal{L}_{\mathrm{effect}}
+
\lambda_r\mathcal{L}_{\mathrm{rel}}
+
\lambda_d\mathcal{L}_{\mathrm{dist}}
+
\lambda_i\mathcal{L}_{\mathrm{inv}}.
\label{eq:eat_objective}
\end{equation}
The loss weights and auxiliary-objective details are given in
Appendix~\ref{app:eat_details}.

After refinement, we freeze the online tokenizer $E_\theta$ and use it
to generate latent-action targets for downstream policy training.
Neither $T_\phi$, the EMA target tokenizer, nor $\Psi$ is required at
policy inference time.
\paragraph{Scope.}
The descriptor $\Psi$ is built from per-cell effects on the pooled feature
grid, and same-source sampling does not guarantee exact geometric
correspondence between donor and recipient states. We therefore interpret EAT
as a source-restricted visual-effect regularizer rather than supervision for a
ground-truth counterfactual physical action.

\begin{algorithm}[t]
\caption{Effect-Anchored Transport (EAT) refinement}
\label{alg:eat}
\begin{algorithmic}[1]
\REQUIRE dataset $\mathcal D$, online tokenizer $E_\theta$, decoder $T_\phi$,
EMA tokenizer $E_{\bar\theta}$, same-source queues $\{Q_d\}$
\FOR{each refinement step}
    \STATE $\bar\theta \leftarrow
    \beta_{\mathrm{EMA}}\bar\theta+(1-\beta_{\mathrm{EMA}})\theta$
    \STATE sample source $d$, donor $x_i$, recipient $F_{j,0}\sim\mathcal D_d$
    \STATE $z_i\leftarrow E_\theta(x_i)$
    \STATE $\hat{D}_{j\leftarrow i}\leftarrow T_\phi(F_{j,0},z_i)$;\;
    $Y_{j\leftarrow i}\leftarrow \sigma_\Delta\hat{D}_{j\leftarrow i}$
    \STATE $\mathcal L_{\mathrm{effect}}\leftarrow
    1-\cos(\Psi(Y_{j\leftarrow i}),
    \operatorname{sg}[\Psi(\Delta F_i)])$
    \STATE $\bar z_i\leftarrow
    \operatorname{sg}[E_{\bar\theta}(x_i)]$;
    enqueue $\bar z_i$ into $Q_d$
    \STATE sample $\bar z_a,\bar z_b\sim Q_d$ and
    $F_{u,0},F_{v,0}\sim\mathcal D_d$
    \STATE $e_s^q\leftarrow
    \Psi(\sigma_\Delta T_\phi(F_{s,0},\bar z_q))$,
    $q\in\{a,b\}$, $s\in\{u,v\}$
    \STATE $\mathcal L_{\mathrm{rel}}\leftarrow
    1-\cos(e_u^a-e_u^b,e_v^a-e_v^b)$
    \STATE compute $\mathcal L_{\mathrm{self}},
    \mathcal L_{\mathrm{dist}},\mathcal L_{\mathrm{inv}}$
    \STATE update $(\theta,\phi)$ with $\mathcal L_{\mathrm{EAT}}$
\ENDFOR
\RETURN $E_\theta$
\end{algorithmic}
\end{algorithm}

\section{Experiments}
\label{sec:experiments}

\subsection{Experimental Setup}
\label{sec:exp_setup}

\paragraph{Pretraining and baselines.}
We train latent-action models on an Open X-Embodiment
mixture~\citep{oxe2024} containing 29 sources and approximately 8.19M clips.
All representation comparisons use the same frozen DINOv2 feature cache. For
the system comparison, TERRA and UniVLA are retrained at the same
latent-action pretraining data scale and evaluated on
LIBERO~\citep{liu2023libero}. For representation analysis we additionally
include a LAPA-style discrete VQ baseline in the same DINO feature space.

\paragraph{Policy protocol.}
Both systems start from OpenVLA-7B~\citep{kim2025openvla}, undergo latent-VLM
pretraining on the same 10\% subset of the pretraining mixture, and are then
fine-tuned on LIBERO.
TERRA uses its continuous query/readout and flow-based policy interface, and
UniVLA its native discrete latent-token interface, so the comparison is
between complete latent-action systems at matched pretraining scale. Each
LIBERO suite contains 10 tasks and is evaluated with 50 rollouts per task.

\paragraph{Representation and robustness evaluation.}
Every tokenizer is frozen and probed with linear readers~\citep{alain2017probes}
fit on real held-out transitions only, following the linear action-decoding
action spaces are never mixed; source-level NMSEs are macro-averaged.
Readers predict five targets from the $K$-step action sequence $a$: the
12-step chunk (\emph{action}), the window sum $N(a)$ (\emph{net}), the first
step (\emph{first}), the profile $A_1(a)$ of Eq.~\ref{eq:temporal_effects}
(\emph{profile}), and the per-step sequence (\emph{seq.}). For nuisance
robustness, the trajectory and labels are fixed while a moving visual
distractor is scaled by $\alpha\in\{0,0.25,0.5,1.0\}$; \emph{nuisance drop}
is the relative loss of net-action explained variance ($1-\mathrm{NMSE}$) at
$\alpha=1$. Donor--recipient distance is the cosine distance between
spatially averaged frozen-DINO initial-state features, ranked within each
source into equal-count quartiles.

\subsection{Closed-loop Manipulation}
\label{sec:closed_loop}

We first evaluate the complete system in closed-loop manipulation.
Table~\ref{tab:closed_loop_main} reports published systems for context and
highlights the retrained TERRA--UniVLA comparison.

\begin{table}[t]
\centering
\includegraphics[width=0.96\linewidth]{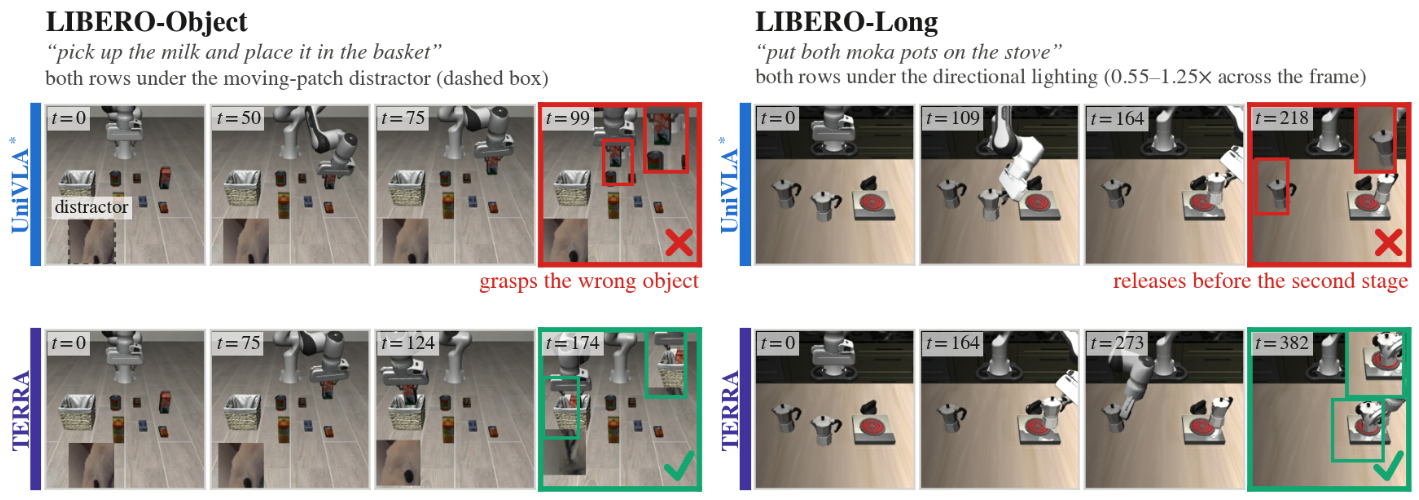}\\[6pt]
\footnotesize
\setlength{\tabcolsep}{15pt}
\begin{tabular}{lccccc}
\toprule
Method & Spatial & Object & Goal & Long & Avg. \\
\midrule
LAPA$^\dagger$             & 73.8 & 74.6 & 58.8 & 55.4 & 65.7 \\
Diffusion Policy$^\dagger$ & 78.3 & 92.5 & 68.3 & 50.5 & 72.4 \\
Octo$^\dagger$             & 78.9 & 85.7 & 84.6 & 51.1 & 75.1 \\
OpenVLA$^\dagger$          & 84.7 & 88.4 & 79.2 & 53.7 & 76.5 \\
\midrule
UniVLA$^*$         & 92.3 & 92.6 & 93.4 & 88.7 & 91.8 \\
\rowcolor{blue!6}
\textbf{TERRA (ours)}      & \textbf{94.1} & \textbf{95.2} & \textbf{93.8} & \textbf{90.6} & \textbf{93.4} \\
\bottomrule
\end{tabular}
\caption{\textbf{Closed-loop LIBERO success rate (\%).}Top: one perturbed rollout each from LIBERO-Object and LIBERO-Long    (Appendix~\ref{app:vla_protocol}). TERRA and UniVLA$^*$ are retrained at the same
latent-action pretraining data scale from the same OpenVLA-7B backbone and
evaluated with the same rollout budget. $^\dagger$Published results under
their original protocols.}
\label{tab:closed_loop_main}
\end{table}

At matched latent-action pretraining scale, TERRA reaches 93.4\% average
LIBERO success against 91.8\% for retrained UniVLA, with higher success on
all four suites; the largest gains are on Object (+2.6) and Long (+1.9),
the suites with the most object and horizon variation. Both systems share
the VLM backbone, the latent-VLM pretraining mixture, and the LIBERO
fine-tuning data; they differ in the latent-action model and the policy
interface built on it. Sections~\ref{sec:representation_results}
and~\ref{sec:eat_results} examine the latent representation directly. The rollouts in Table~\ref{tab:closed_loop_main} illustrate two characteristic
UniVLA$^*$ failures under visual perturbation, distractor confusion on Object
and premature release on Long, which mirror the nuisance-robustness and
within-window-structure results of Section~\ref{sec:representation_results}.
\begin{figure}[t]
    \centering
    \includegraphics[width=0.84\linewidth]{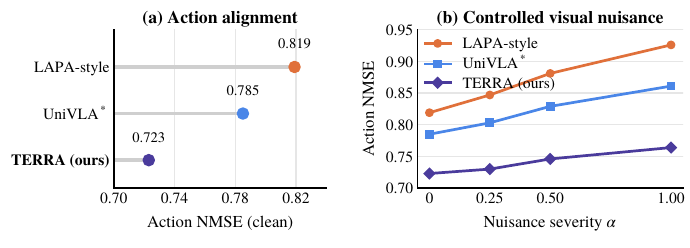}
    \vspace{-2mm}
    \caption{\textbf{Action alignment and nuisance robustness.}
    \textbf{(a)} Action NMSE on clean transitions for TERRA, UniVLA$^*$, and a
    LAPA-style discrete baseline adapted to the common DINO feature space.
    \textbf{(b)} The same reader under a moving visual distractor of severity
    $\alpha$; the robot trajectory and action labels are unchanged.}
    \label{fig:action_alignment}
    \vspace{2mm}
\end{figure}

\subsection{Temporal Effect Representation}
\label{sec:representation_results}

\paragraph{Action alignment and nuisance robustness.}
Figure~\ref{fig:action_alignment} and Table~\ref{tab:ablation_main} evaluate
every tokenizer with the same frozen linear readers. On clean transitions,
TERRA obtains the lowest 12-step action NMSE (.723), against .785 for
UniVLA$^*$ and .819 for the LAPA-style baseline, and the ordering persists as
the distractor severity grows: from $\alpha=0$ to $\alpha=1$, NMSE rises by
.041 for TERRA, .076 for UniVLA$^*$, and .107 for the LAPA-style baseline.
Against UniVLA$^*$, TERRA reads the net action, the first step, and the
per-step sequence more accurately and loses less under the distractor
(17.9\% vs.\ 23.0\%); UniVLA$^*$ reads the action profile slightly better
(.875 vs.\ .909). Profile NMSE stays above .87 for every latent, so the
within-window trend is only weakly linearly decodable from any of them.

\begin{table}[b]
\caption{\textbf{Frozen-reader ablation (NMSE, lower is better).}
UniVLA$^*$ is the retrained baseline of Table~\ref{tab:closed_loop_main}.
The two 65k tokenizers share the architecture and the self-reconstruction
schedule and differ only in whether $A_1$ is provided; the three $+$15k
rows continue from the $[F_0\mid N\mid A_1]$ checkpoint under equal budget.
\emph{Action} predicts the 12-step chunk; \emph{Net}, \emph{First},
\emph{Profile}, \emph{Seq.} predict the net action, the first step,
$A_1(a)$, and the per-step sequence; \emph{Drop} is the relative loss of
net-action explained variance at $\alpha=1$. 95\% intervals are bootstrap
over the 29 sources (Appendix~\ref{app:temporal_diagnostics}). Best per
column underlined.}
\label{tab:ablation_main}
\centering
\footnotesize
\setlength{\tabcolsep}{6pt}
\begin{tabular}{lcccccc}
\toprule
Tokenizer & Action & Net & First & Profile & Seq. & Drop \\
\midrule
UniVLA$^*$                         & .785 & .763 & .778 & \underline{.875} & .785 & 23.0\% \\
\midrule
$[F_0\mid N]$, 65k                 & .817 & .703 & .790 & .983 & .800 & 33.6\% \\
$[F_0\mid N\mid A_1]$, 65k         & .807 & .734 & .767 & .909 & .807 & 22.4\% \\
\quad $+$15k reconstruction        & .784   & .732 & .760 & .909 & .807 & 22.6\% \\
\quad $+$15k EAT w/o invariance    & .764   & .700 & .733 & .915 & .784 & 19.7\% \\
\quad $+$15k full EAT (\textbf{TERRA}) & \underline{.723} & \underline{.671} & \underline{.717} & .909 & \underline{.764} & \underline{17.9\%} \\

\bottomrule
\end{tabular}
\vspace{-2mm}
\end{table}

\paragraph{What the dynamics component adds.}
The two 65k rows of Table~\ref{tab:ablation_main} isolate the temporal
evidence: two tokenizers with the same architecture and schedule, one
receiving the net component only, i.e.\ the frame-pair feature difference
of continuous latent-motion models~\citep{yang2026como}, the other
additionally receiving $A_1$. With the net component alone, the action
profile is not linearly recoverable (.983): the endpoint difference records
how much the scene changed, not how the change was distributed over the
window. Adding $A_1$ lowers profile NMSE to .909, action NMSE from .817 to
.807, first-step NMSE from .790 to .767, and the nuisance drop from 33.6\%
to 22.4\%, while net-action NMSE moves from .703 to .734: the latent trades
a small amount of endpoint precision for within-window structure, and the
added temporal evidence lowers rather than raises sensitivity to the
distractor.

\subsection{Cross-context Refinement with EAT}
\label{sec:eat_results}
\paragraph{Same-budget refinement.}
The $+$15k rows of Table~\ref{tab:ablation_main} hold the optimization
budget fixed. From the shared 65k checkpoint, 15k further
reconstruction-only updates leave the temporal readouts where they were and
lower 12-step action NMSE only from .807 to .784, whereas the same 15k
updates with EAT lower net-action NMSE to .671, first-step NMSE to .717,
sequence NMSE to .764, and action NMSE to .723, and reduce the nuisance
drop to 17.9\%; nuisance invariance accounts for roughly half of each gain.
Profile NMSE is unchanged (.909), consistent with EAT acting on how the
latent behaves across states rather than on the temporal evidence it
encodes, and the net-action readout ends below the endpoint-only tokenizer
(.671 vs.\ .703): the endpoint precision traded for $A_1$ is returned by
EAT, not by additional reconstruction.

\paragraph{Transport.}
To evaluate transport, we fit a linear reader that predicts the recorded
action chunk from a real feature-increment sequence, one per data source,
and freeze it. We then feed it transported transitions $Y_{j\leftarrow i}$
and score the prediction against the donor's action with the same NMSE as
Sec.~\ref{sec:representation_results}. The reader never sees generated
transitions, so a low NMSE means the transported transition still carries
the donor's action the way a real transition would
(details in Appendix~\ref{app:transport_metric}).

\begin{figure}[t]
    \centering
    \includegraphics[width=0.84\linewidth]{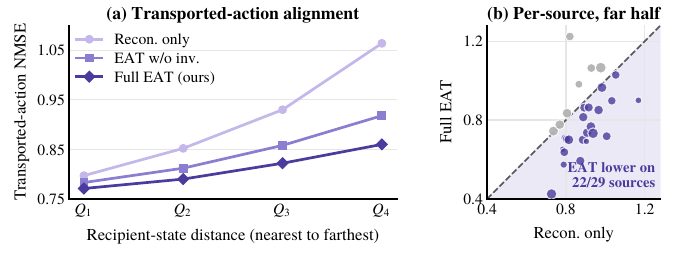}
    \vspace{-2mm}
    \caption{\textbf{Cross-context refinement with EAT.}
\textbf{(a)} Donor-action NMSE of transported transitions
$Y_{j\leftarrow i}$, scored by a reader that is fit only on real
transitions and never sees generated ones, so a low value means the
transported transition still carries the donor's action. Donor--recipient
pairs are grouped by initial-state distance (within-source frozen-DINO
cosine quartiles, $Q_1$ nearest, $Q_4$ farthest). \textbf{(b)} The same NMSE per pretraining source, averaged
over the two farthest quartiles, for reconstruction-only vs.\ full EAT;
each point is one of the 29 sources; points below the diagonal favor EAT
(22 of 29), and gray points mark the sources where EAT is worse.}
    \label{fig:eat_transport}
    \vspace{-1mm}
\end{figure}

EAT makes transport substantially more stable as the recipient context moves
away from the donor. From the nearest to the farthest quartile, donor-action
NMSE rises by .267 under reconstruction-only training, by .136 with EAT
without nuisance invariance, and by .087 with full EAT, a three-fold
reduction in distance sensitivity (Fig.~\ref{fig:eat_transport}a). The
advantage grows with recipient distance, which is exactly the regime EAT
targets, and it holds source by source: over the two farthest quartiles,
full EAT lowers NMSE on 22 of 29 sources (Fig.~\ref{fig:eat_transport}b). The largest degradation is on roboturk (.822 to 1.224). Recipient-state sensitivity and further transport diagnostics
are reported in Appendix~\ref{app:transport_analysis}.

\vspace{-2mm}

\section{Discussion and Limitations}
\label{sec:discussion}

Our transport objective and metric are defined within a data source, where
embodiment and camera geometry are shared; anchoring effects across
embodiments would require a correspondence-aware descriptor. The effect space
keeps only two low-order temporal components on a coarse spatial grid, which
favors robustness over fine temporal detail. The transport metric measures
whether donor action semantics survive decoding in another state, not whether
the decoded transition matches a unique counterfactual future. Finally,
closed-loop results are reported on LIBERO at reduced pretraining scale; larger-scale pretraining and real-robot deployment
remain future work.

\section{Conclusion}

We introduced TERRA, which uses one compact effect space both to define what
a latent action keeps and to supervise how it behaves in new contexts.
Temporal Effect Representation adds a within-window dynamics component to the
endpoint difference, and Effect-Anchored Transport anchors the effect of a
latent decoded in other states to the one observed at its source. The
resulting latents align more closely with actions, degrade more slowly under
visual distractors, and preserve donor actions under transport, and the
complete system improves LIBERO success over UniVLA at matched pretraining
scale.
\newpage
\subsection*{AI use statement}

In this work, we used generative AI tools for drafting, code assistance, experiment orchestration, and language editing.
All AI-assisted code and experimental changes were reviewed and tested by the authors, and all quantitative claims were checked against experiment logs and saved artifacts.
We take responsibility for the final content of this work, including text, claims, and artifacts produced with the aid of generative AI.

\bibliography{iclr2027_conference}
\bibliographystyle{iclr2027_conference}
\appendix


\appendix

\section{Method Details}
\label{app:details}

\subsection{Temporal Effect Representation}
\label{app:temporal_details}

\paragraph{Structured tokens.}
The main tokenizer uses five observations, giving $K=4$ adjacent feature
increments. Frozen DINOv2 features have $C=768$ channels and are pooled to a
$4\times4$ grid ($S=16$). The structured tokens therefore contain 48 tokens:
16 each from $F_0$, the net component $N(\Delta F)$, and the dynamics
component $A_1(\Delta F)$, distinguished by spatial-position and
evidence-type embeddings. A six-layer Transformer aggregates them through
four learned latent queries whose outputs are projected to 64 dimensions,
giving $z\in\mathbb{R}^{4\times 64}$. The tokenizer queries are distinct from
the readout queries used during downstream VLA training and share no
parameters with them.

\paragraph{Temporal-evidence basis.}
For $K=4$, the first nonconstant DCT-II direction used by the dynamics
component has relative coefficients
\[
b_1\propto(0.924,\;0.383,\;-0.383,\;-0.924).
\]
The same coefficient convention is used when constructing the structured
tokens and when evaluating the effect descriptor.

\paragraph{Feature-increment normalization.}
All feature increments are normalized by one fixed global scalar
$\sigma_\Delta$, the root-mean-square magnitude of raw DINOv2 feature
increments, estimated once from real training transitions and reused
throughout training and evaluation. The decoder predicts normalized
increments,
\[
\hat D = T_\phi(F_0,z)\approx \Delta F/\sigma_\Delta ,
\]
and transported transitions are converted back to feature units as
$Y=\sigma_\Delta \hat D$ before the temporal evidence is evaluated on them.

\subsection{Effect-Anchored Transport}
\label{app:eat_details}

\paragraph{Effect descriptor.}
The effect vector $\xi(Y)=\operatorname{vec}[N(Y),A_1(Y)]$ has
$2SC=2\times16\times768=24{,}576$ dimensions. The descriptor first applies a
fixed Gaussian random projection $R\in\mathbb{R}^{24576\times512}$ with
entries drawn i.i.d.\ from $\mathcal N(0,1/512)$ (one draw at
initialization with a fixed seed; no gradient), and then standardizes each
of the $d=512$ output coordinates:
\[
\Psi(Y)=\frac{R^{\top}\xi(Y)-\mu}{\sqrt{\sigma^2}+\epsilon},
\qquad \epsilon=10^{-5},
\]
where $\mu,\sigma^2\in\mathbb{R}^{512}$ are the per-coordinate mean and
variance of $R^{\top}\xi(\Delta F)$ over real training transitions,
estimated once before EAT refinement and then frozen. The standardization
is per coordinate (diagonal), not a full whitening: it equalizes the scale
of the projected coordinates and does not decorrelate them. The same
transform is applied to real and generated transitions, and $\Psi$ has no
learnable parameters.

Because $\xi$ is defined cell by cell on the pooled feature grid and $R$ is
a distance-preserving Johnson--Lindenstrauss
projection~\citep{johnson1984extensions}, matching $\Psi$ constrains the
per-cell image-space effect of the transported transition up to the
projection distortion and the diagonal rescaling; it does not imply
equality of a state-independent physical action.

\paragraph{EMA target tokenizer and donor queue.}
EAT maintains an exponential-moving-average target tokenizer
$E_{\bar\theta}$, initialized from the online tokenizer and updated at the
beginning of each refinement iteration by
\[
\bar\theta\leftarrow
\beta_{\mathrm{EMA}}\bar\theta+(1-\beta_{\mathrm{EMA}})\theta,
\qquad
\beta_{\mathrm{EMA}}=0.99 .
\]
Relative-effect consistency uses detached target latents
$\bar z=\operatorname{sg}[E_{\bar\theta}(x)]$ stored in a per-source FIFO
queue $Q_d$ holding the most recent 16 minibatches (approximately 1024
latents at batch size 64). Because the sampled donor latents are detached,
$\mathcal L_{\mathrm{rel}}$ regularizes the shared decoder directly and
reaches the online tokenizer only indirectly through the jointly optimized
decoder.

\paragraph{Transition-distribution matching.}
Let $y_p=\chi(Y_{j\leftarrow i})$, $p=1,\ldots,m$, denote projected generated
increments in a minibatch and $r_q=\chi(\Delta F_i)$, $q=1,\ldots,n$,
projected real increments, where $\chi$ is a fixed preprocessing and
projection applied identically to both. We use the unbiased squared MMD
\[
\mathcal L_{\mathrm{dist}}
=
\frac{1}{m(m-1)}\sum_{p\neq q}\kappa(y_p,y_q)
+
\frac{1}{n(n-1)}\sum_{p\neq q}\kappa(r_p,r_q)
-
\frac{2}{mn}\sum_{p,q}\kappa(y_p,r_q),
\]
where $\kappa$ is a mixture of RBF kernels. This term matches transition
marginals and does not provide a target counterfactual trajectory for an
individual recipient.

\paragraph{Nuisance invariance.}
The online tokenizer receives a temporally consistent perturbed view while
the EMA tokenizer receives the corresponding clean clip. The perturbation
consists of a square random crop retaining 90\% of the image area (resized
back to the input resolution), brightness scaling sampled from
$\mathcal U(0.8,1.2)$, and contrast scaling sampled from $\mathcal U(0.8,1.2)$.
One augmentation sample is shared by all frames in a clip; no flips,
rotations, hue/saturation changes, or per-frame noise are applied. For a
$224\times224$ input, the crop permits at most about 5\% translation along
either axis, roughly $0.2$ of one cell of the $4\times4$ grid. The augmented
view is used only by $\mathcal L_{\mathrm{inv}}$; $\Psi$ and both
effect-based objectives are evaluated on unaugmented transitions.

\paragraph{Loss weights and schedule.}
The reported refinement objective uses
\[
\lambda_e=1,\qquad
\lambda_r=0.5,\qquad
\lambda_d=1,\qquad
\lambda_i=1 .
\]
The representation is first trained for 65k updates with self-reconstruction
only and then refined for 15k updates with the full EAT objective while
self-reconstruction is retained. After refinement, only the online tokenizer
is retained for generating downstream latent-action targets.

\paragraph{Compute.}
All training uses NVIDIA H100 GPUs. Tokenizer training runs on one GPU:
the 65k self-reconstruction stage takes about 7.8\,h and the 15k EAT stage
about 1.5\,h. Latent-VLM pretraining takes about 18\,h, and LIBERO
fine-tuning takes 5.4--5.6\,h per suite on four GPUs; both are identical for
TERRA and UniVLA$^*$.

\section{Experimental Protocol}
\label{app:evaluation}

\subsection{Baselines}
\label{app:lapa_style}

\paragraph{LAPA-style discrete baseline.}
The representation experiments use a \emph{LAPA-style} discrete VQ baseline,
not the released LAPA tokenizer. We adapt the LAPA-style
bottleneck/reconstruction formulation to the same frozen DINO feature space
used by TERRA and UniVLA and evaluate its discrete codes with the same frozen
linear-reader protocol. The baseline uses NSVQ quantization with a codebook of 8 entries and 4
codes per transition (12 bits), batch size 64, and is trained for 65k
updates on the same data, matching the TERRA self-reconstruction stage; its
codes are one-hot concatenated before the linear reader.

\paragraph{Retrained UniVLA.}
UniVLA is retrained at the same latent-action pretraining data scale as
TERRA. Its tokenizer follows the official recipe, which inherits the released
``Stage-1'' task-irrelevant prior; see the provenance note below.

\subsection{Downstream VLA Protocol and Provenance}
\label{app:vla_protocol}

The closed-loop comparison includes latent-VLM pretraining before LIBERO
fine-tuning. TERRA and retrained UniVLA both start from OpenVLA-7B and use
the same 10\% subset of the latent-action pretraining mixture for this stage.
All closed-loop numbers use a single fine-tuning seed, 10 tasks per suite,
and 50 rollouts per task.

TERRA uses a continuous policy interface: four learned readout embeddings
are placed after the image and instruction, and their hidden states condition
a flow-matching expert that jointly predicts the continuous latent target and
a 12-step, 7-DoF action chunk. UniVLA uses its native discrete latent-token
interface. Table~\ref{tab:closed_loop_main} therefore compares complete
latent-action systems at matched pretraining scale rather than fixing the
policy interface.

At inference, TERRA uses only the current observation and language
instruction. Future frames, the TERRA tokenizer, the EMA tokenizer, the
conditional decoder, and the effect descriptor are not executed online.

\paragraph{Provenance of the retrained UniVLA row.}
\label{app:provenance}
The retrained UniVLA row in Table~\ref{tab:closed_loop_main} is not the
published UniVLA model. Published UniVLA uses its original larger-scale
pretraining protocol, whereas the 91.8\% row uses the reduced 10\%
latent-VLM pretraining mixture described above. The retrained recipe also
inherits the small released ``Stage-1'' task-irrelevant prior required by the
official recipe. We therefore report it as a matched-scale system baseline
rather than a strict same-provenance-from-random comparison.

\paragraph{Closed-loop perturbation.}
\label{app:perturb}
The rollouts shown above Table~\ref{tab:closed_loop_main} are evaluated
under the LIBERO-Plus perturbations~\citep{fei2025liberoplus} at
level L3 (lighting, background texture, and object layout with distractor
objects), plus the moving distractor of Appendix~\ref{app:metrics}, on the
10 tasks of LIBERO-Object and LIBERO-Long with 50 rollouts per task (500
episodes per suite and condition). Perturbed rollouts are used for the
qualitative comparison only; the success rates in
Table~\ref{tab:closed_loop_main} are unperturbed.

\subsection{Action Readers and Controlled Nuisance}
\label{app:metrics}

All representation comparisons freeze the tokenizer. Because the 29 OXE
sources use heterogeneous action spaces, we fit a separate linear action
reader within each source $d$ and macro-average the source-level normalized
mean-squared error
\[
\mathrm{NMSE}_d
=
\frac{\sum_i\|a_i-\hat a_i\|_2^2}
     {\sum_i\|a_i-\bar a_d\|_2^2},
\]
where $\bar a_d$ is the mean action of source $d$. Readers are ridge
regressors. Held-out transitions are split by episode, two thirds for
fitting and one third for scoring; the ridge coefficient is selected from
$\{10^{-1},1,10,10^{2},10^{3},10^{4}\}$ on a 20\% episode-level validation
split of the fitting portion. The same split and protocol are used for all
tokenizers and all five targets of Sec.~\ref{sec:exp_setup}.

For nuisance robustness, the underlying robot trajectory and action labels
are held fixed while a moving visual distractor is scaled by
$\alpha\in\{0,0.25,0.5,1.0\}$. The distractor is a $64\times64$ natural-video patch pasted opaquely onto
the $224\times224$ frame; it moves along the frame border (inset 32\,px) at
4--8\,px per frame, with its trajectory and speed determined by the episode
id. The severity $\alpha$ scales the distractor magnitude: $\alpha=0$ is
the clean frame and $\alpha=1$ the fully pasted patch. The same distractor
realization is reused across severity levels; only its magnitude changes.

\subsection{Transported-Action Evaluation and Context Sensitivity}
\label{app:transport_metric}

\paragraph{Transported-action reader.}
To evaluate transport without training on generated samples, we use a
\emph{transported-action reader}
$g_{\mathrm{act}}:\mathbb{R}^{K\times S\times C}\to\mathbb{R}^{12\times 7}$,
a linear map from a feature-increment sequence to the 12-step action chunk
used in Sec.~\ref{sec:representation_results}. One reader is fit per data
source by ridge regression on \emph{real} held-out pairs
$(\Delta F_i/\sigma_\Delta,\,a_i)$ and then frozen. For a transported
transition we score
\[
\hat a_i=g_{\mathrm{act}}(Y_{j\leftarrow i}/\sigma_\Delta)
\]
against the recorded donor action $a_i$ with the source-wise NMSE of
Sec.~\ref{sec:exp_setup}. Because $g_{\mathrm{act}}$ never sees generated
transitions and reads the increment sequence directly rather than
$\Psi(Y)$, a low NMSE means the transported transition carries the donor's
action in the same form a real transition would.

\paragraph{Recipient distance.}
Recipient distance is computed independently of any learned tokenizer. Let
\[
c_i=
\operatorname{norm}
\left(
\frac{1}{S}\sum_{s=1}^S F_{i,0}^{(s)}
\right),
\qquad
d_{\mathrm{ctx}}(i,j)=1-c_i^\top c_j ,
\]
i.e.\ the cosine distance between spatially averaged frozen DINO
initial-state features. Distances are ranked within each data source and
divided into equal-count quartiles; $Q_1$ contains the nearest 25\% of
donor--recipient pairs and $Q_4$ the farthest 25\%.

\paragraph{Context sensitivity.}
To monitor whether the decoder ignores the recipient state, we report
\[
\mathrm{CtxSens}
=
\frac{
\|\Psi(T_\phi(F_{j,0},z))-\Psi(T_\phi(F_{\ell,0},z))\|_2
}{
\|\Psi(T_\phi(F_{j,0},z))\|_2+\epsilon
},
\]
averaged over same-source recipient pairs $(j,\ell)$. A value near zero
would indicate little recipient dependence in the descriptor space.

\section{Additional Results}
\label{app:additional_results}

\subsection{Bootstrap Intervals for Table~\ref{tab:ablation_main}}
\label{app:temporal_diagnostics}

\begin{table}[t]
\caption{\textbf{95\% bootstrap intervals} for the readouts of
Table~\ref{tab:ablation_main}, obtained by resampling the 29 sources
2{,}000 times. Intervals reflect variation across sources, not across
training seeds (tokenizer training uses one seed).}
\label{tab:ablation_ci}
\centering
\scriptsize
\setlength{\tabcolsep}{3.5pt}
\begin{tabular}{lccccc}
\toprule
Tokenizer & Net & First & Profile & Seq. & Drop (\%) \\
\midrule
$[F_0\mid N]$, 65k              & .703 [.644, .762] & .790 [.712, .864] & .983 [.964, 1.000] & .800 [.730, .866] & 33.6 [24.2, 46.9] \\
$[F_0\mid N\mid A_1]$, 65k      & .734 [.653, .813] & .767 [.700, .834] & .909 [.875, .942] & .807 [.741, .866] & 22.4 [13.7, 33.2] \\
\quad $+$15k reconstruction     & .732 [.650, .815] & .760 [.692, .828] & .909 [.874, .943] & .807 [.740, .869] & 22.6 [14.5, 33.4] \\
\quad $+$15k EAT w/o invariance & .700 [.615, .781] & .733 [.663, .806] & .915 [.879, .951] & .784 [.715, .848] & 19.7 [12.6, 27.4] \\
\quad $+$15k full EAT           & .671 [.584, .759] & .717 [.647, .790] & .909 [.878, .940] & .764 [.693, .831] & 17.9 [11.1, 24.9] \\
\midrule
UniVLA$^*$                      & .763 [.703, .819] & .778 [.729, .824] & .875 [.836, .911] & .784 [.729, .833] & 23.0 [12.5, 34.1] \\
\bottomrule
\end{tabular}
\end{table}
The same-budget comparison isolates EAT refinement from additional
reconstruction training and isolates the nuisance-invariance term; it does
not individually isolate $\mathcal L_{\mathrm{effect}}$,
$\mathcal L_{\mathrm{rel}}$, and $\mathcal L_{\mathrm{dist}}$.

\subsection{Transport Diagnostics}
\label{app:transport_analysis}

\begin{table}[t]
\caption{\textbf{Cross-context transport diagnostics.} Context sensitivity
is $\mathrm{CtxSens}$ from Appendix~\ref{app:transport_metric}; latent
effective rank is $\exp(-\sum_i p_i\log p_i)$ with $p_i=\sigma_i/\sum_j
\sigma_j$ over the singular values $\sigma_i$ of the centered
$4\times64=256$-dimensional latent matrix on held-out
transitions~\citep{roy2007effective}.}
\label{tab:transport_diag}
\centering
\footnotesize
\begin{tabular}{lccc}
\toprule
Diagnostic & 65k start & EAT w/o invariance & Full EAT \\
\midrule
Context sensitivity       & --    & .465 & .487 \\
Latent effective rank     & 210.9 & --   & 165.7 \\
\bottomrule
\end{tabular}
\end{table}

The nonzero context-sensitivity values show that changing the recipient
state materially changes the transported descriptor. The full model's value
of $.487$ argues against a state-independent donor replay as the sole
explanation for its lower transported-action NMSE.

\begin{table}[t]
\caption{\textbf{Per-source transported-action NMSE over the two farthest
recipient quartiles} (data of Fig.~\ref{fig:eat_transport}b). $n$ is the
number of held-out test clips per source. Full EAT is lower on 22 of 29
sources; the mean over sources is .893 (reconstruction only) vs.\ .806
(full EAT).}
\label{tab:per_source_far}
\centering
\scriptsize
\setlength{\tabcolsep}{4pt}
\begin{tabular}{lrcc@{\hspace{14pt}}lrcc}
\toprule
Source & $n$ & Recon. & EAT & Source & $n$ & Recon. & EAT \\
\midrule
austin\_buds & 84 & .905 & .693 & fractal20220817\_data & 120 & .875 & .593 \\
austin\_sailor & 114 & 1.009 & .718 & furniture\_bench & 123 & .808 & .835 \\
austin\_sirius & 144 & .790 & .644 & iamlab\_cmu\_pickup\_insert & 109 & .931 & 1.065 \\
bc\_z & 128 & .929 & .767 & jaco\_play & 124 & .801 & .708 \\
berkeley\_autolab\_ur5 & 140 & .729 & .425 & kuka & 126 & .917 & .865 \\
berkeley\_cable\_routing & 126 & .968 & .851 & language\_table & 116 & .895 & .863 \\
berkeley\_fanuc\_manipulation & 120 & .888 & .699 & nyu\_franka\_play & 130 & .817 & .701 \\
berkeley\_gnm\_cory\_hall & 80 & 1.170 & .900 & roboturk & 105 & .822 & 1.224 \\
berkeley\_gnm\_recon & 128 & .985 & .966 & stanford\_hydra & 113 & .794 & .637 \\
berkeley\_gnm\_sac\_son & 109 & 1.055 & 1.030 & taco\_play & 123 & .890 & .815 \\
bridge\_big & 128 & .739 & .744 & toto & 108 & 1.035 & .898 \\
cmu\_stretch & 126 & .910 & .736 & ucsd\_kitchen & 88 & .791 & .574 \\
dlr\_edan\_shared\_control & 95 & .868 & .982 & utaustin\_mutex & 122 & .770 & .777 \\
dobbe & 141 & .979 & 1.067 & viola & 142 & .940 & .733 \\
fmb & 115 & .897 & .869 & & & & \\
\bottomrule
\end{tabular}
\end{table}

\subsection{Spatial Evidence Resolution}
\label{app:spatial_resolution}

We evaluate spatial evidence resolution under a fixed representation budget.
Frozen DINO patch features are pooled to different grids, the same
$(N,A_1)$ temporal evidence is constructed, and every representation is
PCA-compressed to 256 dimensions before applying the same linear reader.
No tokenizer is retrained for this sweep.

\begin{table}[t]
\caption{\textbf{Spatial evidence resolution under matched 256-D
compression} (net-action NMSE, lower is better). Nuisance drop is defined
in Sec.~\ref{sec:exp_setup}.}
\label{tab:spatial_res}
\centering
\footnotesize
\begin{tabular}{ccccc}
\toprule
Grid & Cells & Clean & Distractor & Nuisance drop \\
\midrule
$16\times16$ & 256 & .675 & .977 & 93.0\% \\
$8\times8$   & 64  & .658 & .860 & 59.0\% \\
$4\times4$   & 16  & .642 & .799 & 44.0\% \\
$2\times2$   & 4   & .650 & .810 & 45.7\% \\
\bottomrule
\end{tabular}
\end{table}

Under this matched compression/readout protocol, robustness improves
strongly as resolution is reduced to $4\times4$ and then largely plateaus.
Because finer grids undergo stronger PCA compression, this result does not
imply that a learned full-patch tokenizer is intrinsically worse; it shows
that additional spatial capacity admits nuisance faster than it improves
linear action readout under a fixed readout budget.

\end{document}